\documentclass{article}
\usepackage{ijcai25}

\usepackage{times}
\usepackage{soul}
\usepackage{url}
\usepackage{color}
\usepackage[hidelinks]{hyperref}
\usepackage[utf8]{inputenc}
\usepackage[small]{caption}
\usepackage{pict2e}
\usepackage{arydshln}
\usepackage{graphicx}
\usepackage{amsmath}
\usepackage{amsthm}
\usepackage{booktabs}
\usepackage{xcolor}
\usepackage{subcaption}
\usepackage{bbm}
\usepackage{amssymb}
\usepackage[linesnumbered,ruled]{algorithm2e}
\usepackage[switch]{lineno}
\usepackage{dsfont}
\def\method{\texttt{HGEN}}

\newtheorem{theorem}{Theorem}
\newcommand{\cupdot}{\mathbin{\mathaccent\cdot\cup}}

\newtheorem{remark}{Remark}

\title{HGEN: Heterogeneous Graph Ensemble Networks}

\author{
Jiajun Shen$^1$
\and
Yufei Jin$^1$\and
Kaibu Feng$^2$\and
Yi He$^2$\And
Xingquan Zhu$^1$\\
\affiliations
$^1$Dept. of Electrical Engineering and Computer Science, Florida Atlantic University, USA\\
$^2$Department of Data Science, William \& Mary,  USA\\
\emails
\{jshen2024, yjin2021, xzhu3\}@fau.edu;
\{kfeng03, yihe\}@wm.edu
}

\begin{document}

\maketitle

\begin{abstract}

This paper presents \method\ that pioneers
ensemble learning for heterogeneous graphs.
We argue that the heterogeneity in node types, nodal features, 
and local neighborhood topology poses significant challenges for ensemble learning, particularly in accommodating diverse graph learners.
Our \method\ framework ensembles multiple learners through a meta-path and transformation-based optimization pipeline to uplift classification accuracy. 
Specifically, \method\ uses meta-path combined with random dropping to create \textit{Allele Graph Neural Networks (GNNs)}, whereby the base graph learners are trained and aligned for later ensembling. 
To ensure effective ensemble learning,
\method\ presents two key components:
1) a \textit{residual-attention} mechanism to calibrate allele GNNs of different meta-paths, thereby enforcing node embeddings to focus on more informative graphs to improve base learner accuracy, and 
2) a \textit{correlation-regularization} term to enlarge the disparity among 
embedding matrices generated from different meta-paths, 
thereby enriching base learner diversity.
We analyze the convergence of  \method\ and attest its 
higher regularization magnitude over simple voting.
Experiments on five heterogeneous networks validate that \method\ consistently outperforms 
its state-of-the-art competitors by substantial margin. 
Codes are available at \url{https://github.com/Chrisshen12/HGEN}.

\end{abstract}

\section{Introduction}

Ensemble learning that combines multiple base models to enhance predictive performance has achieved remarkable success across diverse domains, from weather forecasting~\cite{Molteni1996ECMWF} and online trading~\cite{Sun2023stock} to image classification~\cite{Yang2023Fewshot} and recent prompt-based ensembling in Large Language Models~\cite{Zhang2024AAAI}. While individual models often struggle with generalization and overfitting, ensemble learning harnesses the diversity of base models to improve robustness, reduce variance, and achieve superior accuracy.


Despite its wide adoption, ensemble learning has been mostly studied in the context of \textit{i.i.d.} data, leaving its application to data with complex interdependencies such as graphs relatively underexplored. 
%
%
Early studies such as graph representation ensembling~\cite{9381465} fused node embeddings from multiple graph learning models through concatenation. Subsequent methods such as stacking-based frameworks~\cite{CHEN2022113753} proposed multi-level classifiers to aggregate representations for tasks like link prediction. More recent efforts such as Graph Ensemble Neural Network (GEN)~\cite{DUAN2024102461} integrated ensemble learning directly within Graph Neural Networks (GNNs), performing ensemble operations throughout the training process rather 
than solely at the prediction stage.
To further improve robustness and mitigate overfitting and adversarial attacks, recent methods such as Graph Ensemble Learning (GEL)~\cite{math10081300} introduced serialized knowledge passing and multilayer DropNode strategies to promote diversity, while GNN-Ensemble \cite{ensemblegnn} employed substructure-based training to defend against adversarial perturbations.

While these methods have advanced ensemble learning for homogeneous graphs, 
they falter in dealing with {\bfseries heterogeneous graphs}, which are commonly observed in real-world applications, such as social networks \cite{social}, citation networks \cite{citation}, urban networks \cite{urban}, and biomedical networks \cite{Jin2024HGDL}. Indeed, ensemble learning on heterogeneous graphs has three unique challenges as follows.
1) {\em Graph heterogeneity}, where diverse node and edge types 
necessitate specialized techniques such as meta-path~\cite{10.1145/3292500.3330961} or attention-based~\cite{han2019} models to extract meaningful relationships and patterns,
2) {\em Base learner accuracy}, where the non-uniform structure of heterogeneous graphs requires  accurate base learners to ensure reliable ensemble performance, and 
3) {\em Base learner diversity}, where ensuring diverse learners is crucial for 
robust and generalized predictions, given the varied nature of heterogeneous graphs.

\begin{figure*}[htbp]
    \centering
    \includegraphics[width=0.85\textwidth]{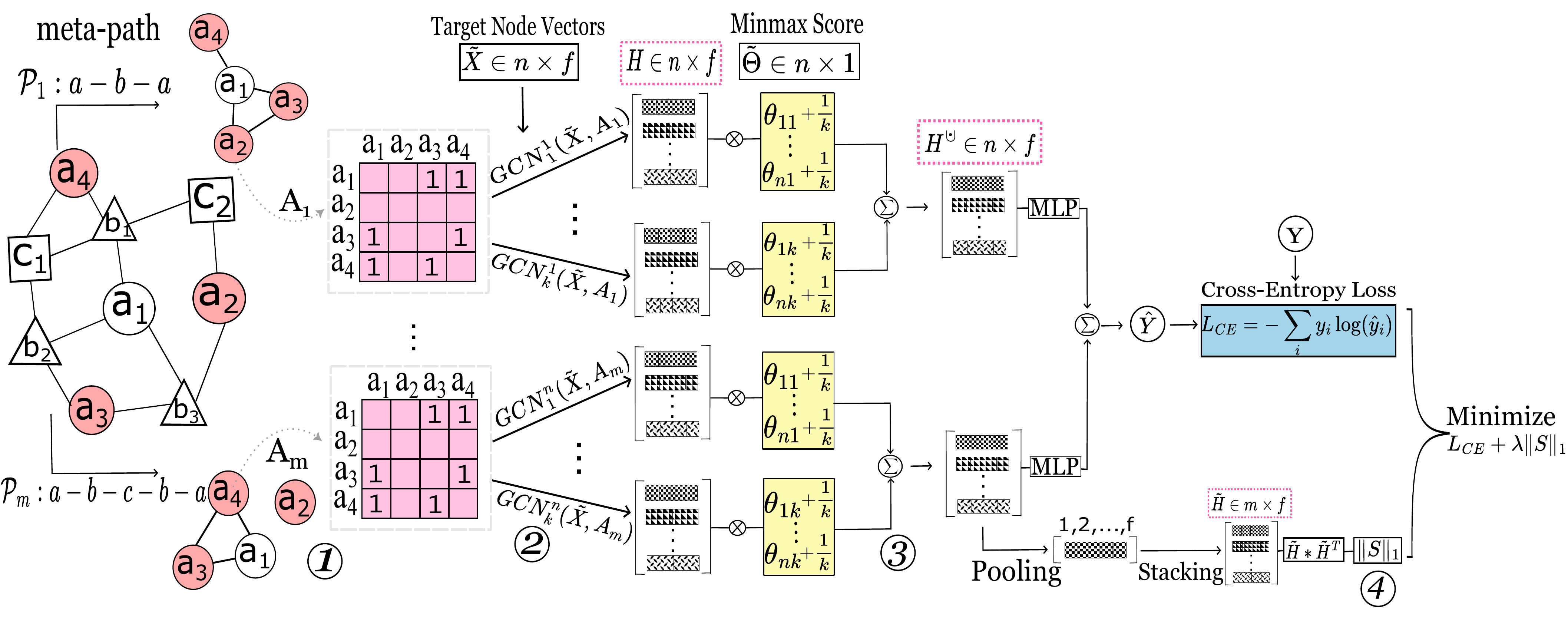}
    \caption{The proposed \method\ for heterogeneous graph ensemble learning. From left to right, \textcircled{1}: a heterogeneous graph is first converted to $m$ meta-graphs (one for each meta-path), with $A_m$ denotes adjacency matrix of the graph from the $m^{th}$ meta-path; \textcircled{2}: node feature dropout is applied to each meta-graph and help train $k$ GNN learners (\textit{i.e.} Allele GNNs); \textcircled{3}: Residual-attention is applied to allele GNNs of each meta-path to consolidate their node embedding features, with one multi-layer-perceptron (MLP) project layer is learned from each meta-path; and \textcircled{4} a correlation regularizer enforces meta-path's embedding features to be different from each other and ensembles of MLPs to have minimum cross-entropy loss. The combined objective function enforces the GNNs, residual-attention, and MLP project layers to collectively learn for optimized ensemble learning goal.}
    \label{fig:framework}
\end{figure*}

To address these challenges, we propose \method,
a novel ensemble learning framework with solid theoretical foundation for heterogeneous graphs.
\method\ is tailored to accommodate various mainstream graph learners, 
such as Graph Convolutional Networks (GCNs), Graph Attention Networks (GATs), and GraphSAGE,
offering a flexible and efficient solution for complex networked data.
Specifically, \method\ simultaneously enhances learner accuracy and diversity
by devising a regularized allele GNN framework, 
which employs feature and edge dropping to generate diverse GNNs from heterogeneous graphs.
This enhances generalization by encouraging different perspectives of the graph for promoting diversity across base learners. To further improve accuracy, a residual-attention mechanism is incorporated to enable adaptive ensemble weighting, allowing base learners to aggregate allele GNNs 
derived from perturbed graph structures while favoring more informative graphs.
%
%
Unlike existing serialized or boosting-based ensemble methods, 
which can be computationally expensive hence impractical for large graphs,
\method\ adopts a bagging strategy, 
enabling a balance between computational efficiency and scalability
through parallelization.

\smallskip \noindent
{\bf Specific contributions} of this paper includes the following:
\begin{enumerate}
    \item \method\ is the first framework to enable ensemble learning for heterogeneous graphs, tackling the unique challenges posed by their complexity and diversity. 

    \item A novel attention-based aggregation that dynamically adjusts weights based on the contributions of base graph learners is proposed. A residual attention mechanism is to refine ensemble weighting further, enhancing adaptability and improving overall predictive accuracy.

    \item Our theoretical analysis substantiates the 
    convergence of \method, and that its magnitude of correlation regularization overweighs na\"ive voting.

    \item Empirical study on five real-world heterogeneous networks 
validates the effectiveness of \method\ over the recent arts, where the datasets establish a new benchmark for heterogeneous graph ensemble learning.
\end{enumerate}
\section{Problem Definition}

Let $G=(V,E,X,Y)$ denote a heterogeneous graph, where $V$ is the set of nodes, 
$E$ is the set of edges, 
$X$ is the feature matrix for the nodes of a certain targeted type, 
and $Y$ is the one hot encoding label matrix for those target nodes. 
Denote $\mathcal{T}^v$ as the set of node types and $\mathcal{T}^e$ as the set of edge types, where $t_{i} \in \mathcal{T}^v$ is the node type $i$ and $e_{i,j} \in \mathcal{T}^e$ is the edge type that connects the node type from $t_{i}$ to $t_{j}$. 

Define $\phi:~V \mapsto \mathcal{T}^v$ that maps the node from $V$ to its node type, and $\varphi:~E \mapsto \mathcal{T}^e$ that maps the edges from $E$ to its edge type. A  meta-path $\mathcal{P}$ is a relational sequence $(e_{i,j}e_{j,k}\ldots e_{k,j}e_{j,i})$ that is symmetric. 
Given a specific meta-path $\mathcal{P}_{i}$, we can construct a homogeneous graph equipped with an adjacency matrix $A_{i} \in \mathds{1}^{n \times n}$ where $A_{i}[j, k]=1$ $\iff$ $\exists~p = (v_{j},\dots, v_{k})$ such that edges in the path $p$ match the meta-path $\mathcal{P}_i$, formally presented as follows.

\begin{equation}
\begin{aligned}
\text{Gen}(G, \mathcal{P}) &= A_i \quad \text{where}\\ 
A_i[j,k] &=
\begin{cases} 
1 & \text{if } \exists p = (v_j, \dots, v_k) \text{ matching } \mathcal{P}_i, \\
0 & \text{otherwise}.
\end{cases}
\end{aligned} \label{eq:path}
\end{equation}
Given graph $G$ and a target node type $t_{\iota}\in\mathcal{T}^v$, with $V_{t_{\iota}} \subset V$ being the target node set containing $n$  target nodes, a simple graph learner such as GNN can only learn from a homogeneous graph.
Our \textbf{goal} is to enable ensemble learning of multiple GNNs to predict the label $Y$ for the target node set $V_{t_{\iota}}$ so to maximize the classification accuracy and AUC values.

\section{Proposed Framework}

The proposed \method\ ensemble learning framework, in Figure~\ref{fig:framework}, uses meta-paths to extract multiple disparate homogeneous graphs from the heterogeneous graph,
such that base graph learners like GCN, GAT, or GraphSAGE can be trained from the extracted graphs to form an ensemble.
Two major components of \method\ to enhance base learner accuracy and diversity
are presented in Sections~\ref{subsec:accuracy} and~\ref{subsec:diversity}, respectively,
followed by the analysis.
Main steps of \method\ are outlined in Algorithm 1 in Appendix.


\subsection{\method\ Base Learner Accuracy Enhancement}
\label{subsec:accuracy}


\subsubsection{Allele Graph Neural Network Learning}
For each meta-path $\mathcal{P}_{i}$, we extract a homogeneous graph $G_i=(A_i, X_i)$ consisting of target nodes and its adjacency matrix denoted by $A_i$ and nodal features $X_i$. Although GNNs under certain conditions such as Lipschitz graph filters are stable~\cite{Gama2020stability}, most GNNs intend have unstable learning outcomes, especially when trained with limited samples with random initialization. 
This motivates us to devise \textit{allele graph neural networks} for ensemble learning, where alleles refer to variant GNN networks learned from the same source. Therefore, we propose to augment each meta-path graph by using node feature dropout, as follows.
\begin{align}
    \tilde{X}_i &= \text{Dropout}(X_i,b)\\
    H_i^{(0)} &= \sigma(\tilde{X}_i W_i^{0})
\end{align}
where $b$ is the dropout rate, $\sigma(\cdot)$ is the non-linear activation function, $X^i$ is the original node features, $W_i^{0}$ is the projection learnable parameters, and $H_i^{(0)}$ is the embedding output prepared for graph learning. Instead of using more sophisticated graph augmentation approaches, our ablation study in Sec.~\ref{sec:experiment} will soon demonstrate that feature dropout outperforms other alternatives. This is also consistent with previous observations where feature dropout often outperforms node or edge dropout~\cite{Shu2022Dropout}. 


After applying base graph learner GNN to augmented meta-path graph $\tilde{G}_i=(A_i, \tilde{X}_i)$, we can obtain one base learner. For each meta-path $\mathcal{P}_{i}$, we apply different initializations and dropout to obtain $k$ base GNN models. With $m$ meta-paths, a total of $(k*m)$ base GNNs are created. Each single base GNN consists of a single linear projection layer projecting raw feature input $X$ along and multiple graph filtering layers aggregating embeddings with corresponding adjacency matrix $A_{i}$. The projection layer is a simple linear layer projecting raw features with a random feature dropout followed by a nonlinear-activation. For the message passing scheme, we used three different backbones: graph convolution layer, graph attention layer, and GraphSAGE layer to show the effects of base model variants. In general, any standard message passing scheme can fit into the framework. 

In the following, we outline the $j^{th}$ GNN from the $i^{th}$ meta-path $\mathcal{P}_{i}$ using GCN, and the layer-wise aggregation of the node information is as follows:
\begin{align}
    H_{i,j}^{(l)} = \sigma(D_i^{-\frac{1}{2}}(A_i+I)D_i^{-\frac{1}{2}}H_{i,j}^{(l-1)}W_{i,j}^{(l)})
\end{align}
with $D_i$ denoting the degree matrix of $(A_i+I)$, $W_{i,j}^{(l)}$ as the learnable parameters for the convolution layer $l$. With $L$ graph convolution layers in total $l\in\{1,\cdots,L\}$, the GCN outputs of the final node embeddings for the $j^{th}$ GNN from the $i$th meta-path $\mathcal{P}_{i}$ is represented as $H_{i,j}^{(L)}$.

\subsection{Intra Fusion for each Meta-Path}
\begin{figure}[htbp]
    \centering
    \includegraphics[width=0.45\textwidth]{{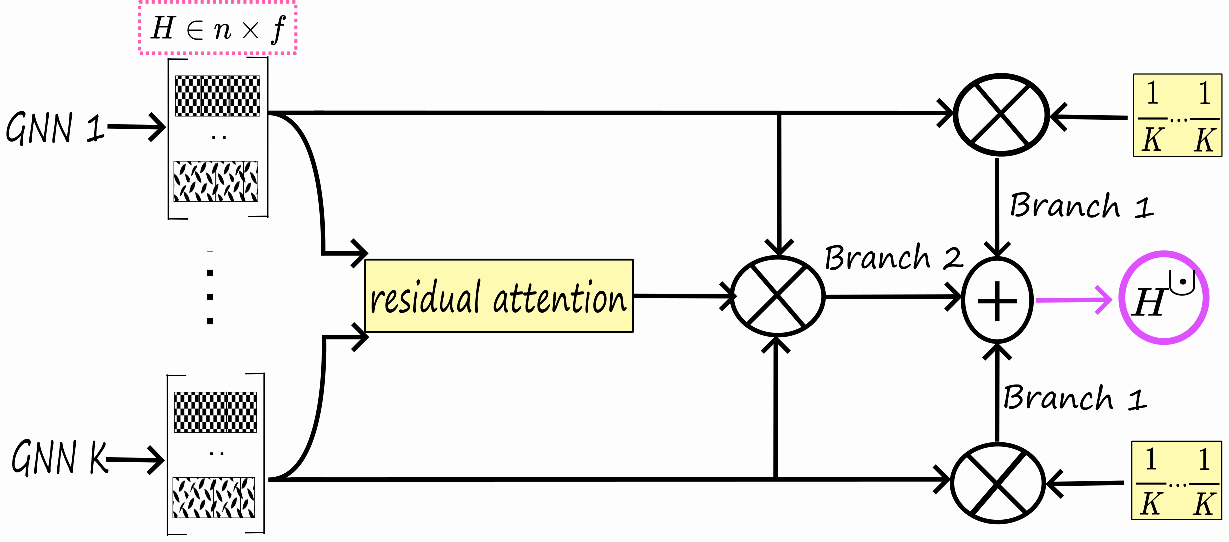}}
    \caption{Residual attention concept: instead of learning the attention for each GCN with only one branch, we use the residual mechanism to learn a ``Branch 2'' representing perturbation deviated from the ``Branch 1'' which is a simple average ensemble GNN.}
    \label{fig:attention-concept}
\end{figure}
\begin{figure}[htbp]
    \centering
    \includegraphics[width=0.45\textwidth]{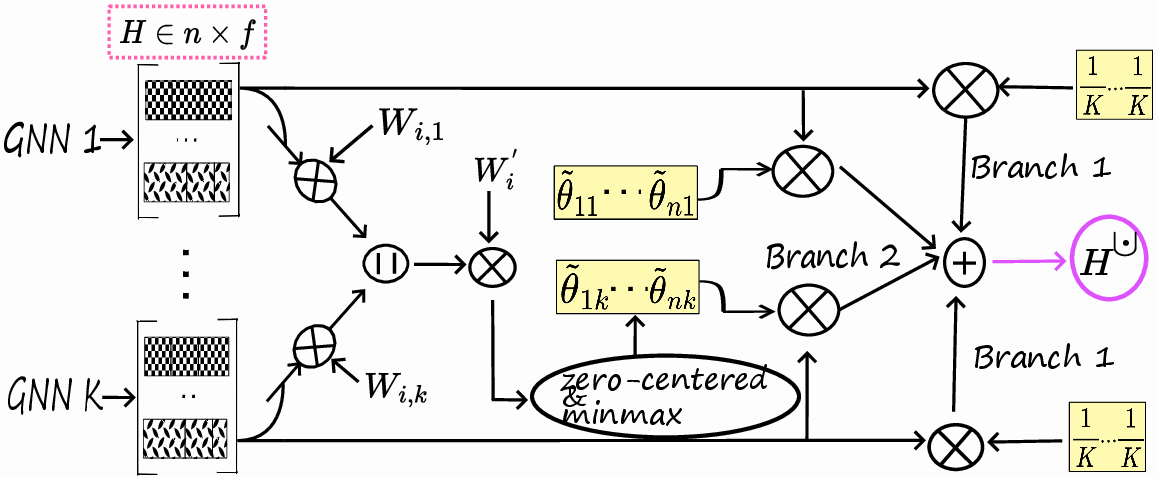}
    \caption{Residual attention computation: $k$ GNNs first being compressed to attention space through projection weight $W_{i,j}$ and then being concatenated and fed to a shared projection weight $W'_{i}$ to learn $k$ $\Theta$ attention scores. Using minmax normalization (refer to Eq~\ref{eq:attention_org}), the final fusion residual attention $\tilde{\Theta}$ is obtained. Final fusion representation $H^{\cupdot}$ is the summation of ``Branch 1'' and ``Branch 2''.}
    \label{fig:attention-detail}
\end{figure}

\subsubsection{Residual-Attention for Allele GNN Fusion}
The allele GNNs each learns from an augmented meta-path graph, conveying unique information of the underlying meta-path semantics and the network heterogeneity. As the first key step of the ensemble learning process, a fusion mechanism is introduced to consolidate embedding features from allele GNNs to adaptively adjust weight of respective GNNs.

Instead of directly learning GNN weights, we introduce a residual-attention mechanism which borrows residual network's~\cite{He2015} unique strength of learning identity mapping that is hard to directly learn from highly nonlinear functions. The residual mechanism provides a good precondition or initial points that guarantee easy learning on recovering shallow layer results. Meanwhile, it is known that the attention mechanism is highly non-linear and could be difficult to learn a uniform weight distribution. Inspired by the residual mechanism and its success in image recognition~\cite{wang2017residual}, we propose a residual-attention approach for the first ensemble stage that ensures an easy learning of allele GNN aggregation, which is shown to be a simple mechanism serving as a good precondition analogy to identity mapping.

Specifically,
for each meta-path $\mathcal{P}_i$, our model leverages multiple single GNN to learn the embeddings. To aggregate the final embeddings of multiple GNNs, a residual-attention based fusion mechanism is designed as follows:
\begin{align}
\Theta &= \|_{j=1}^{k}
 \big\{ (H^{(L)}_{i,j}) W_{i,j} \big\}
 W_i^{'}, \quad \text{and} \label{eq:attention_org}\\ 
 \bar{\Theta} &= \text{Mean}(\Theta), ~~ \hat{\Theta} = \Theta-\bar{\Theta}, ~~
 \tilde{\Theta} = 
 \frac{(\hat{\Theta}-\hat{\Theta}^{\downarrow})} {(\hat{\Theta}^{\uparrow}-\hat{\Theta}^{\downarrow})}
 \nonumber
\end{align}

where $\Theta \in \mathbb{R}^{n \times k} $ is the learned residual attention with meta-path $\mathcal{P}_i$, where $n$ denotes number of target nodes and $k$ denotes the number of allele GNNs for each meta-path. $\bar{\Theta}$ is the row average of learned $\Theta$. $\hat{\Theta}$ is the broadcasting difference between $\Theta$ and $\bar{\Theta}$, which is zero-centered. ${\Theta}^{\uparrow}= \max (\hat{\Theta})$ and ${\Theta}^{\downarrow}= \min (\hat{\Theta})$ are the corresponding maximal and minimal attention scores over row. $\tilde{\Theta}$ is the final minimax version of residual attention. \textcolor{black}{The min-max residual attention adaptively adjusts the weights/influence of each base learner. Strong learners making more correct predictions are promoted with higher weights, while weaker learners have lower weights, allowing the model to mitigate the impact of errors without explicit error detection.}

\begin{figure*}
    \centering
    \includegraphics[width=1\linewidth]{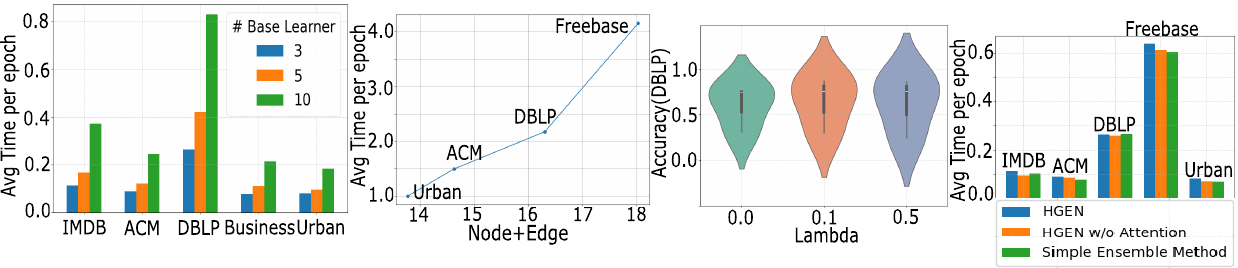}
    \caption{(a) Average runtime per epoch \textit{w.r.t.} different number of base learners on five datasets. (b) Log Average runtime per epoch using three base learners increases with number of nodes plus number of edges in log term. (c) Ensemble accuracy of each meta-path \textit{w.r.t.} different $\lambda$ values. (d) Runtime comparison on HGEN, HGEN w/o attention, and simple ensemble model on five datasets, where Freebase is the largest to attest their scalability.}
    \label{fig:one_row}
\end{figure*} 


\paragraph{Residual-Attention Fusion.} In order to fuse embeddings from allele GNNs which are derived from the same meta-path graph, we employ a residual-attention mechanisms, as defined in Eq.~(\ref{attention_H}), where $\tilde{\Theta}[:,i]+\frac{1}{k}$ is the node-wise attention and $+$ is the broadcasting plus over nodes as follows.
\begin{align}    
    H_i^{\cupdot} &= \sum_{j=1}^{k}(\tilde{\Theta}[:,j]+\frac{1}{k}) \cdot H^{(L)}_{i,j} \label{attention_H}
\end{align}
As the results of the residual-attention, $H_i^{\cupdot}$ is used to denote the aggregated node embeddings for meta-path $\mathcal{P}_i$ after residual-attention fusion. It is worth noting that the residual-attention in Eq.~(\ref{attention_H}) does not involve any learnable parameters. The attention in Eq.~(\ref{eq:attention_org}) learns respective parameters to regulate the residual-attention fusion. 


\subsection{\method\ Base Learner Diversity Enhancement}
\label{subsec:diversity}

After obtaining the final embedding $H^{\cupdot}_{i}$ for each meta-path $\mathcal{P}_{i}$, we project the embedding to the output class dimension with a linear layer decoder.
Finally, we use summation to bag the prediction for each meta-path to obtain final prediction:
\begin{equation}
    \hat{Y} = \sum_{i=1}^m \text{MLP}(H^{\cupdot}_i),
\end{equation}
where a mean pooling operator $\text{MP}(\cdot)$ is used to obtain graph embeddings $\tilde{H}_{i} \in R^{f}$ over $H^{\cupdot}_{i}$ for each meta-path $\mathcal{P}_{i}$. Denote the stacked version of all meta-paths $\tilde{H}_{i}$ as $\tilde{H} \in \mathbb{R}^{m \times f}$. A correlation matrix is computed quantifying the inter correlation among meta-paths:
\begin{equation}
    S = \tilde{H}*\tilde{H}^T
\end{equation}

where $S \in \mathbb{R}^{m \times m}$ is the correlation matrix for evaluating inter meta-path relation. We applied $L_1$ norm on $S$ as regularization loss to ensure more independent embeddings learned among all meta-paths. The final objective function of the \method\ is then defined by Eq.~(\ref{eq:objective}).

\begin{equation}
    \ell= - \sum_{i} y_i \log(\hat{y}_i) + \lambda \| S \|_1
    \label{eq:objective}
\end{equation}

\textcolor{black}{We evaluated the impact of correlation regularization term by varying its weight coefficient $\lambda$ as indicated in Eq.~(\ref{eq:objective}).
A larger $\lambda$ enforces greater sparsity in the correlation matrix, encouraging base learners trained from different meta-path graphs to be more independent.
To assess this effect, we examined the ensemble performance across different meta-paths for $\lambda \in \{0, 0.1, 0.5\}$, as shown in Fig. ~\ref{fig:one_row}(c) . 
The violin plots visualize the distribution of ensemble performance. As $\lambda$ increases, the spread (\textit{i.e.}, width) of the distributions becomes larger, suggesting that models trained from individual meta-paths produce increasingly diverse predictions. This validates that the correlation-regularization term positively contributes to diversity between ensemble models.}


\subsection{Theoretical Analysis}

\paragraph{Complexity.}
Algorithm 1 in Appendix lists the major steps of \method~which takes a heterogeneous graph as input, learns allele GNNs from meta-path graphs, and outputs predictions for target nodes. 
Given $m$ meta-paths and $k$ allele GNNs for each meta-path, with an average of $e$ number of edges for each meta-path. Denote training a single GNN time complexity is $T$ with $T$ at least linearly scaled to $\mathcal{O}(nf+e)$. The asymptotic complexity of the \method~is then at least $\mathcal{O}(m*(kT)+m*(k*nf)$. With $m*k*nf$ roughly the time for the residual attention computation and $m*(kT)$ for each individual GNN training. Since $nf\leq T$, we should approximately have $c*m*(kT)$ time complexity and asymptotically $\mathcal{O}(m*(kT))$ time complexity. With the assumption that $m,k \ll T$, the total complexity of the ensemble model is linearly scalable.

\textcolor{black}{We report the wallclock runtime performance by assessing the average training time per epoch across ensemble sizes in Fig.~\ref{fig:one_row}(b) .
We observe a linear increase of runtime \textit{w.r.t.} ensemble size in all five datasets.
Besides, on larger graphs, the increasing trend of average training time per epoch 
\textit{w.r.t.} increasing number of nodes and edges is also linear.
The two results suggest scalability of HGEN in both ensemble size and graph size, attesting its practicability for large-scale graph learning. 
We also conduct experiments by using a large Freebase dataset containing 40,000 nodes and four different magnitude datasets, with the results illustrated in Fig.~\ref{fig:one_row}(d) .
%
%
We observe that the attention in HGEN could introduce slight overhead, but it remains constant and does not scale up with the graph size.
If we remove this attention mechanism, the runtime of HGEN will be on a par with the naive ensemble baseline.
The extra runtime is mainly required from computing the attention, it will remain manageable for large graphs.}

\paragraph{Convergence \& Superiority Than Naive Voting.} In the following, we derive a Theorem and two remarks, which assert the convergence of \method, and its superiority compared to simple voting based ensemble. Detailed derivations are reported in Supplement D.
\begin{theorem}

Denoted by $L'= -\sum_{i} y_i \log(y_i) $ the cross-entropy loss accumulated by by minimizing the objective Eq.~(\ref{eq:objective}) over $t$ training iterations.  The cumulative loss $\ell$ at the $t$-th iteration satisfies:
\begin{align*}
\ell &= L' + \| S \|_1 \\
  &\leq t \cdot L'(W_t,W_*^{\text{mlp}}) + \frac{ \|W_t^{mlp}-W_*^{mlp}\|}{2\eta'} \\ & + m \cdot f \cdot \max |h_{ij}^{\cupdot}|^2,
\end{align*}
where $W_t$ and $W_t^{\text{mlp}}$ represent the learnable parameters of base graph learner and the weights of MLP at  the $t$-th iteration, respectively;
$W_*^{\text{mlp}}$ is the optimal MLP weights and $\eta'$ the step size. 
$L$ is the Lipschitz constant.
Denoted by $h_{ij}^{\cupdot}$ the $(i,j)$-th entry of the fused node embedding matrix $H_j^{\cupdot}$, whose magnitude is bounded by:
\begin{equation*}
h_{ij}^{\cupdot} \in 
\begin{aligned}[t]
\Big(
&\min\Big(
        {\mu - 4\sigma - \sqrt{t}\,\eta L}, 
        {(k+1)(\mu - 4\sigma - \sqrt{t}\,\eta L)}
    \Big),\\
&\max\Big(
        {\mu + 4\sigma + \sqrt{t}\,\eta L}, 
        {(k+1)(\mu + 4\sigma + \sqrt{t}\,\eta L)}
    \Big)
\Big).
\end{aligned}
\end{equation*}

\end{theorem}


\begin{remark}
Consider the immediate loss $l_t (W_t)$ that holds 
\resizebox{\columnwidth}{!}{$l_t (W_t) \leq L'(W_t,W_*^{\text{mlp}}) + \frac{ \|W_t^{mlp}-W_*^{mlp}\|}{2\eta't} + \frac{m \cdot f \cdot \max |h_{ij}^{\cupdot}|^2}{t}.$}
We observe 
$\lim_{t \rightarrow \infty} ( L'(W_t,W_*^{\text{mlp}}) - l_t (W_t) ) = 0$, 
where $l_t (W_t)$ 
is approaching its optimal status as $t$ goes larger.
This suggests that the learning process 
of using MLP to ensemble the resultant 
node embeddings of base graph learners
is stabilizing and improving over time,
indicating convergence.


\end{remark}

\begin{remark}
The $\theta$-weighted ensemble strategy in \method\ enjoys a magnified 
term $\| S \|_1$ hence strengthen the regularization effect through the aggregated embedding , because 
$\textnormal{range}(h_{ij}^{\cupdot}) \geq 
\textnormal{range}(\tilde{h}_{ij}^{\cupdot})$,
where $\tilde{h}_{ij}^{\cupdot} \in 
(\mu-4\sigma-\sqrt{t}\eta L,\mu+4\sigma+\sqrt{t}\eta L)$
is the $(i,j)$-th entry of the fused node embedding matrix generated from na\"ive voting, and $\widetilde{\Theta}_{i,j} \in [0,1],~\forall i,j$.
This means that \method\ provides a broader range of possible embedding values compared to the na\"ive voting,
thereby resulting in more flexible and informative embeddings, which helps improve overall performance and generalization.




\end{remark}

\section{Experiment}
\label{sec:experiment}

\subsection{Benchmark Datasets}
Five heterogeneous graphs from real applications are used as benchmark datasets. Their statistics
and detailed descriptions are deferred to Supplement B of Appendix due to page limits. 
\begin{table*}[htbp]

\label{tab:overall result}
\centering
\resizebox{0.9\textwidth}{!}{%
\begin{tabular}{@{}llll|lll|lll@{}}
\toprule
    Base Learner:    &              & GCN                       &                           &               & GraphSAGE                 &                           &              & GAT                       &                           \\ \midrule
Dataset & Model        & Accuracies                       & AUC                       & Model         & Accuracies                       & AUC                       & Model        & Accuracies                       & AUC                       \\ \midrule
IMDB    & HAN$_{GCN}$     & 0.540$_{\pm0.0160}^*$          & 0.720$_{\pm0.0072}^*$          & HAN$_{SAGE}$     & 0.523$_{\pm0.0059}^*$          & 0.697$_{\pm0.0121}$          & HAN$_{GAT}$     & 0.563$_{\pm0.0219}^*$          & 0.747$_{\pm0.0140}^*$          \\
        & SeH$_{GCN}$          & 0.536$_{\pm0.0031}^*$          & 0.712$_{\pm0.008}^*$           & SeH$_{SAGE}$           & 0.398$_{\pm0.0101}^*$          & 0.568$_{\pm0.0117}^*$           & SeH$_{GAT}$          & 0.419$_{\pm0.0101}^*$          & 0.591$_{\pm0.0123}^*$           \\
        & GCN-Ensemble & 0.551$_{\pm0.0406}^*$          & 0.747$_{\pm0.0184}^*$          & SAGE-Ensemble & 0.589$_{\pm0.0055}^*$          & 0.760$_{\pm0.0021}^*$          & GAT-Ensemble & 0.583$_{\pm0.0034}^*$          & 0.754$_{\pm0.0017}^*$          \\
        & NaiveWeighting$_{GCN}$ & 0.596$_{\pm0.0031}^*$          & 0.774$_{\pm0.0029}^*$          & NaiveWeighting$_{SAGE}$ & 0.587$_{\pm0.0067}^*$          & 0.762$_{\pm0.0063}^*$          & NaiveWeighting$_{GAT}$ & 0.595$_{\pm0.0034}$          & 0.770$_{\pm0.0030}$          \\
        & Transformer$_{GCN}$  & 0.595$_{\pm0.0075}^*$          & 0.771$_{\pm0.0034}^*$          & Transformer$_{SAGE}$   & 0.584$_{\pm0.0072}^*$          & 0.757$_{\pm0.0047}^*$          & Transformer$_{GAT}$  & 0.591$_{\pm0.0036}^*$          & 0.767$_{\pm0.0034}$          \\
        & HGEN$_{GCN}$    & \textbf{0.604$_{\pm0.0033}$} & \textbf{0.776$_{\pm0.0010}$} & HGEN$_{SAGE}$    & \textbf{0.605$_{\pm0.0040}$} & \textbf{0.775$_{\pm0.0032}$} & HGEN$_{GAT}$    & \textbf{0.600$_{\pm0.0021}$} & \textbf{0.769$_{\pm0.0034}$} \\ \midrule
ACM     & HAN$_{GCN}$     & 0.839$_{\pm0.0183}^*$          & 0.973$_{\pm0.0015}^*$          & HAN$_{SAGE}$     & 0.880$_{\pm0.0174}^*$          & 0.977$_{\pm0.0032}^*$          & HAN$_{GAT}$     & 0.872$_{\pm0.0107}^*$          & 0.965$_{\pm0.0066}^*$          \\
        & SeH$_{GCN}$          & 0.794$_{\pm0.0168}^*$          & 0.923$_{\pm0.0145}^*$          & SeH$_{SAGE}$           & 0.753$_{\pm0.0213}^*$          & 0.914$_{\pm0.0121}^*$          & SeH$_{GAT}$          & 0.730$_{\pm0.0577}^*$          & 0.898$_{\pm0.0385}^*$          \\
        & GCN-Ensemble & 0.766$_{\pm0.0088}^*$          & 0.969$_{\pm0.0030}^*$          & SAGE-Ensemble & 0.802$_{\pm0.0308}^*$          & 0.984$_{\pm0.0012}$          & GAT-Ensemble & 0.825$_{\pm0.0078}^*$          & 0.978$_{\pm0.0005}$          \\
        & NaiveWeighting$_{GCN}$ & 0.892$_{\pm0.0089}^*$          & 0.977$_{\pm0.0017}$          & NaiveWeighting$_{SAGE}$ & 0.909$_{\pm0.0113}^*$          & 0.984$_{\pm0.0009}$          & NaiveWeighting$_{GAT}$ & 0.892$_{\pm0.0062}$          & 0.978$_{\pm0.0005}$          \\
        & Transformer$_{GCN}$  & 0.898$_{\pm0.0103}^*$          & 0.977$_{\pm0.0019}$          & Transformer$_{SAGE}$   & 0.912$_{\pm0.0017}^*$          & 0.983$_{\pm0.0008}^*$          & Transformer$_{GAT}$  & 0.905$_{\pm0.0041}$          & 0.978$_{\pm0.0007}$          \\
        & HGEN$_{GCN}$    & \textbf{0.909$_{\pm0.0016}$} & \textbf{0.977$_{\pm0.0016}$} & HGEN$_{SAGE}$    & \textbf{0.923$_{\pm0.0022}$} & \textbf{0.984$_{\pm0.0006}$} & HGEN$_{GAT}$    & \textbf{0.908$_{\pm0.0007}$} & \textbf{0.977$_{\pm0.0011}$} \\ \midrule
DBLP    & HAN$_{GCN}$     & 0.868$_{\pm0.0247}^*$          & 0.970$_{\pm0.0097}^*$          & HAN$_{SAGE}$     & 0.891$_{\pm0.0161}^*$          & 0.975$_{\pm0.0048}^*$          & HAN$_{GAT}$     & 0.900$_{\pm0.0100}^*$          & 0.982$_{\pm0.0020}^*$          \\
        & SeH$_{GCN}$          & 0.809$_{\pm0.0175}^*$          & 0.878$_{\pm0.0355}^*$          & SeH$_{SAGE}$           & 0.780$_{\pm0.0266}^*$          & 0.926$_{\pm0.0155}^*$          & SeH$_{GAT}$          & 0.878$_{\pm0.0094}^*$          & 0.971$_{\pm0.0026}^*$          \\
        & GCN-Ensemble & 0.925$_{\pm0.0123}$          & 0.990$_{\pm0.0014}$          & SAGE-Ensemble & 0.930$_{\pm0.0022}^*$          & 0.990$_{\pm0.0006}$          & GAT-Ensemble & 0.867$_{\pm0.0439}^*$          & 0.977$_{\pm0.0046}^*$          \\
        & NaiveWeighting$_{GCN}$ & 0.932$_{\pm0.0017}$          & 0.990$_{\pm0.0007}^*$          & NaiveWeighting$_{SAGE}$ & 0.931$_{\pm0.0021}$          & 0.989$_{\pm0.0007}$          & NaiveWeighting$_{GAT}$ & 0.919$_{\pm0.0071}^*$          & 0.984$_{\pm0.0016}$          \\
        & Transformer$_{GCN}$  & 0.913$_{\pm0.0179}^*$          & 0.948$_{\pm0.0114}^*$          & Transformer$_{SAGE}$   & 0.928$_{\pm0.0025}^*$          & 0.988$_{\pm0.0014}^*$          & Transformer$_{GAT}$  & 0.902$_{\pm0.0103}^*$          & 0.983$_{\pm0.0013}^*$          \\
        & HGEN$_{GCN}$    & \textbf{0.932$_{\pm0.0020}$} & \textbf{0.991$_{\pm0.0003}$} & HGEN$_{SAGE}$    & \textbf{0.936$_{\pm0.0021}$} & \textbf{0.989$_{\pm0.0015}$} & HGEN$_{GAT}$    & \textbf{0.928$_{\pm0.0031}$} & \textbf{0.987$_{\pm0.0018}$} \\ \midrule
Business & HAN$_{GCN}$     & 0.717$_{\pm0.0022}^*$          & 0.782$_{\pm0.0015}^*$          & HAN$_{SAGE}$     & 0.720$_{\pm0.0030}^*$          & 0.779$_{\pm0.0029}^*$          & HAN$_{GAT}$     & 0.692$_{\pm0.0193}^*$          & 0.744$_{\pm0.0296}^*$          \\
        & SeH$_{GCN}$          & 0.702$_{\pm0.0134}^*$          & 0.759$_{\pm0.0163}^*$          & SeH$_{SAGE}$           & 0.678$_{\pm0.0037}^*$          & 0.708$_{\pm0.0093}^*$          & SeH$_{GAT}$          & 0.597$_{\pm0.0927}^*$          & 0.587$_{\pm0.1487}^*$          \\
        & GCN-Ensemble & 0.708$_{\pm0.0012}^*$          & 0.775$_{\pm0.0006}^*$          & SAGE-Ensemble & 0.710$_{\pm0.0026}^*$          & 0.772$_{\pm0.0008}^*$          & GAT-Ensemble & 0.705$_{\pm0.0038}^*$          & 0.774$_{\pm0.0016}^*$          \\
        & NaiveWeighting$_{GCN}$ & 0.715$_{\pm0.0035}^*$          & 0.770$_{\pm0.0329}$          & NaiveWeighting$_{SAGE}$ & 0.720$_{\pm0.0030}^*$          & 0.784$_{\pm0.0022}^*$          & NaiveWeighting$_{GAT}$ & 0.712$_{\pm0.0038}^*$          & 0.778$_{\pm0.0051}^*$          \\
        & Transformer$_{GCN}$  & 0.719$_{\pm0.0023}^*$          & 0.786$_{\pm0.0018}^*$          & Transformer$_{SAGE}$   & 0.721$_{\pm0.0064}^*$          & 0.783$_{\pm0.0045}^*$          & Transformer$_{GAT}$  & 0.713$_{\pm0.0021}^*$          & 0.780$_{\pm0.0021}^*$          \\
        & HGEN$_{GCN}$    & \textbf{0.725$_{\pm0.0042}$} & \textbf{0.788$_{\pm0.0011}$} & HGEN$_{SAGE}$    & \textbf{0.732$_{\pm0.0019}$} & \textbf{0.787$_{\pm0.0018}$} & HGEN$_{GAT}$    & \textbf{0.726$_{\pm0.0059}$} & \textbf{0.785$_{\pm0.0027}$} \\ \midrule
Urban   & HAN$_{GCN}$     & 0.231$_{\pm0.0155}^*$          & 0.596$_{\pm0.0090}^*$          & HAN$_{SAGE}$     & 0.502$_{\pm0.0178}^*$          & 0.811$_{\pm0.0061}^*$          & HAN$_{GAT}$     & 0.368$_{\pm0.0650}^*$          & 0.765$_{\pm0.0470}$          \\
        & SeH$_{GCN}$          & 0.204$_{\pm0.0037}^*$          & 0.458$_{\pm0.0062}^*$          & SeH$_{SAGE}$           & 0.329$_{\pm0.0276}^*$          & 0.779$_{\pm0.0175}^*$          & SeH$_{GAT}$          & 0.206$_{\pm0.0000}^*$          & 0.487$_{\pm0.0637}^*$          \\
        & GCN-Ensemble & 0.206$_{\pm0.0000}^*$           & 0.416$_{\pm0.0026}^*$          & SAGE-Ensemble & 0.454$_{\pm0.0185}^*$          & 0.832$_{\pm0.0052}^*$          & GAT-Ensemble & 0.300$_{\pm0.0207}^*$          & 0.761$_{\pm0.0134}^*$          \\
        & NaiveWeighting$_{GCN}$ & 0.246$_{\pm0.0367}^*$          & 0.532$_{\pm0.0798}^*$          & NaiveWeighting$_{SAGE}$ & 0.538$_{\pm0.0108}^*$          & 0.831$_{\pm0.0125}$          & NaiveWeighting$_{GAT}$ & 0.444$_{\pm0.0578}^*$          & 0.815$_{\pm0.016}$           \\
        & Transformer$_{GCN}$  & 0.201$_{\pm0.0045}^*$          & 0.457$_{\pm0.0034}^*$          & Transformer$_{SAGE}$   & 0.500$_{\pm0.0477}^*$          & 0.815$_{\pm0.0091}^*$          & Transformer$_{GAT}$  & 0.383$_{\pm0.0531}$          & 0.787$_{\pm0.0212}^*$          \\
        & HGEN$_{GCN}$    & \textbf{0.289$_{\pm0.0000}$} & \textbf{0.612$_{\pm0.0090}$} & HGEN$_{SAGE}$    & \textbf{0.591$_{\pm0.0142}$} & \textbf{0.850$_{\pm0.0115}$} & HGEN$_{GAT}$    & \textbf{0.451$_{\pm0.0353}$} & \textbf{0.813$_{\pm0.0106}$} \\ \bottomrule

\end{tabular}%
}
\caption{Performance comparisons between baselines and our proposed method equipped with GCN, GraphSAGE, and GAT graph base learners across five heterogeneous datasets. Accuracies (ACC) and AUC values are reported over 5 different initialization status. Superscript *  indicates that \method\ is statistically significantly better than this method at 95\% confidence level using the performance metrics.}
\label{tab:overall result}
\end{table*}

\subsection{Baselines}
We compare our \method\  with some state-of-art baselines including  heterogeneous graph embedding models.
\begin{itemize}
    \item \textbf{HAN}~\cite{han2019} leverages node- and semantic-level attention mechanisms
    to learn heterogeneous node embeddings from different meta-paths.
\end{itemize}
\begin{itemize}
    \item \textbf{Ensemble-GNN} is a variant of the state-of-the-art ensemble learning method for homogeneous graphs~\cite{ensemblegnn}, which combines predictions from multiple GNNs through voting. To make the method~\cite{ensemblegnn} working for hoterogeneous graphs, we use meta-paths to generate homogenerous graphs, and apply the method~\cite{ensemblegnn} for ensemble learning (predictions are generated using GCN, GraphSAGE, and GAT). After training, the predictions across all GNNs and meta-paths are aggregated through voting to determine the final prediction. 
\end{itemize}
\begin{itemize}
    \item \textbf{Transformer-GNN} uses a transformer architecture to combine embeddings from different GNNs (GCN, GraphSAGE, and GAT) for each meta-path. Once predictions are generated for a meta-path, they are stacked and integrated to produce the final prediction, ensuring an effective fusion of heterogeneous graph information.
\end{itemize}
\begin{itemize}
    \item \textbf{SeHGNN}~\cite{SeH} is a heterogeneous graph representation learning approach that precomputes neighbor aggregation using a lightweight mean aggregator. It extends the receptive field through long meta-paths and integrates information using a transformer-based fusion module.
\end{itemize}
\begin{itemize}
    \item \textbf{NaiveWeighting-GNN} is a variant of the proposed \method. It uses the same architecture and loss function (and regularizer) as \method\ to guide learning but replaces the residual-attention fusion of \method\ using a simple mean average to aggregate all allele GNNs. Its comparision to \method\ can demonstrate the advantage of the residual-attention fusion, compared to simple voting.
\end{itemize}

\subsection{Implementation Details}
We perform a grid search with selected range of hyperparameters including hidden dimension, layer size, dropping rate, number of individual GNN, and control rate for regularizer. We choose Adam~\cite{adam} as our optimizer. We fix the learning rate, weight decay, the number of epochs and apply early stopping mechanism. For each method, we report the average accuracy and roc-auc score across five random seeds. All experiments are run on desktop workstations equipped with Nvidia GeForce RTX 2080 Ti. 

\subsection{Results and Analysis}
\paragraph{Variants and Baseline Comparison.}
Table~\ref{tab:overall result} reports the results of the experiment on five datasets with different baselines, our proposed method, and variants over three individual message passing backbones, including graph convolution network (GCN), graph attention network (GAT), and GraphSAGE. Within the same message passing scheme, it can be observed that our proposed \method~consistently performs better over other baselines with $95\%$ confidence level on IMDB, Business, ACM, and Urban datasets, and scores on top along with the GCN-Ensemble method on DBLP dataset, proving the superiority of our framework. 

Compared to all baselines (excluding NaiveWeighting GNN, which is \method's variant), \method~performs consistently better over all datasets with $95\%$ confidence level. For GCN backbones, \method~beats all other variants over IMDB, Business, ACM, and Urban dataset and performs on top along with weighted GCN over DBLP dataset. For GAT backbones, our methods is on par with Transformer$_{GAT}$ and weighted GAT over Urban and ACM datasets and outperforms others in the rest of datasets. This shows the advantage and necessity of our individual components.

Note, \method\ is statistically significantly better than GNN-Ensemble on 25 out of 30 occasions (across all five datasets). Although both of them are graph ensemble learning methods, GNN-ensemble's base learners are not regulated by the global objective function and there is no constraint to enhance base learner diversities. This results in suboptimal base learners with low accuracies and diversity. 

Comparing \method\ and its variant NaiveWeighting$_{GNN}$, the results show that \method\ is statistically significantly better than NaiveWeighting-GNN on 17 out of 30 occasions (across all five datasets), asserting the advantage of the proposed residual-attention for allele GNN fusions, compared to naive voting. This is also consistent with our theoretical analysis in Section 3.3 which asserts that the $\Theta$-weighted ensemble strategy strengthens the regularization effects due to its larger range of $h_{ij}^{\cupdot}$, compared to naive voting. 

\begin{figure}
    \centering
    \includegraphics[width=0.95\linewidth]{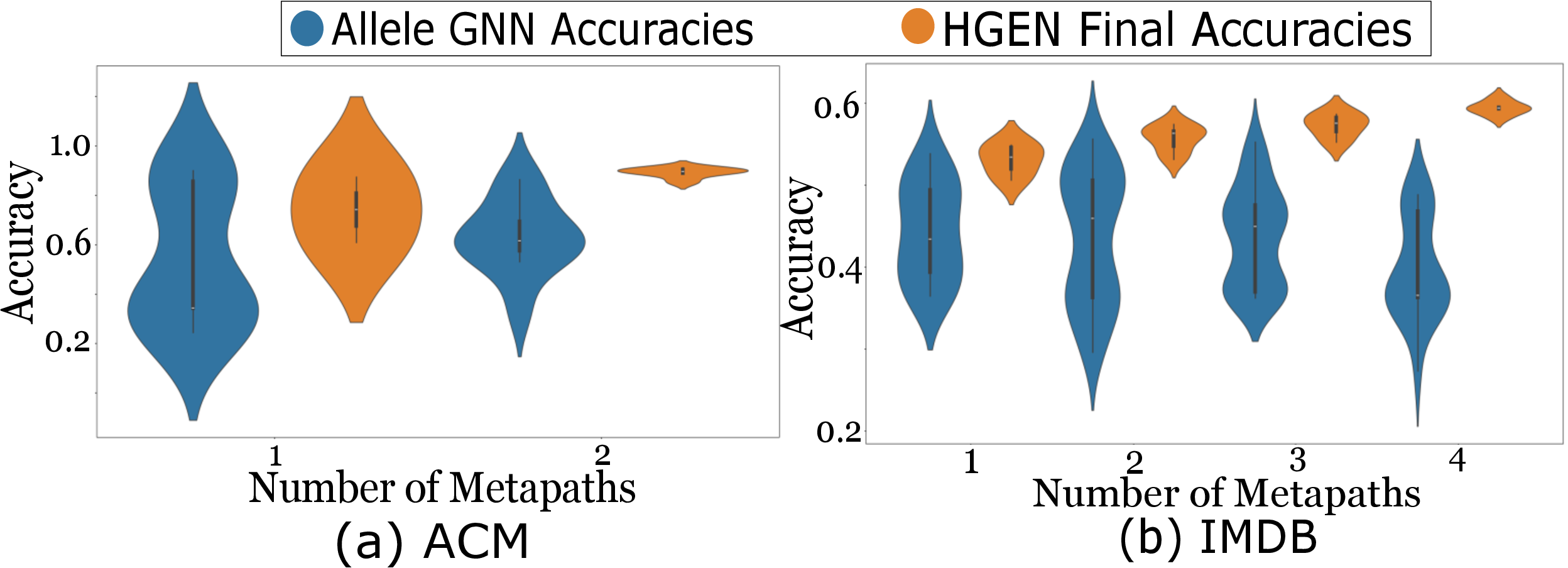}
    \caption{Impact of allele GNNs and meta-paths on the ensemble learning results (using GCN as the base learner). Blue violin plots show allele GNNs' mean and variance whereas orange violin plots show \method's mean and variance.}
    \label{fig:volinVariance}
\end{figure}

\paragraph{Ablation Study on Allele GNNs \& Meta-paths.}
In order to validate the impact of allele GNNs on \method's ensemble learning results, we report mean and variance for allele GNNs leaner and \method\ in Figures~\ref{fig:volinVariance}(a) and~\ref{fig:volinVariance}(b), where the blue violin plots represents allele GNN models' accuracies (mean and variance), and the the orange violin plots show the corresponding final accuracy of \method. For each dataset, the results are reported by increasing the number of meta-paths. Figures~\ref{fig:volinVariance}(a) and~\ref{fig:volinVariance}(b) show that the variance of the blue plots is significantly larger than that of the orange plots. This suggests that individual allele GNN models exhibits larger variability and diversity in their predictions, a preferable setting for ensemble learning. As a result, the final accuracy, after ensemble, shows a more stable and consistent result.

As we add more meta-paths, more GNNs are obtained. Each GNN model learns different aspects of the data, improving the overall robustness of the model. However, while individual GNNs show more diversity, their performance can still be limited by the inherent biases or weaknesses of each model.
Meanwhile, when comparing individual GNN accuracy and \method~final accuracy, we observe that the ensemble results have a much higher accuracy, showing the success of our residual attention ensemble approach. Using attention mechanisms, \method\ learns dynamic weights to each meta-path, focusing on capturing and fusing individual GNN's advantage in the learning process. By enforcing diversity across models from different meta-paths through the $\| S \|_1$, \method\ strengthens their embedding fusion $\textnormal{range}(h_{ij}^{\cupdot})$ as we have theoretically analyzed in Section 3.3, and therefore improves the final prediction results.


\begin{table}[htbp]

\centering
\resizebox{\columnwidth}{!}{%
\begin{tabular}{llll}
\hline
Dataset & Model                                   & ACC              & AUC              \\ \hline
IMDB    & HGEN feature drop+regularizer           & \textbf{0.605}$_{\pm0.0040}$ & \textbf{0.775}$_{\pm0.0032}$ \\
        & HGEN feature drop                       & 0.600$_{\pm0.0112}$ & 0.773$_{\pm0.0071}$ \\
        & HGEN edge drop+regularizer              & 0.582$_{\pm0.0143}$ & 0.755$_{\pm0.0168}$ \\
        & HGEN feature drop+edge drop+regularizer & 0.595$_{\pm0.0034}$ & 0.771$_{\pm0.0050}$ \\ \hline
ACM     & HGEN feature drop+regularizer           & \textbf{0.923}$_{\pm0.0022}$ & \textbf{0.984}$_{\pm0.0006}$ \\
        & HGEN feature drop                       & 0.909$_{\pm0.0083}$ & 0.983$_{\pm0.0014}$ \\
        & HGEN edge drop+regularizer              & 0.914$_{\pm0.0067}$ & 0.983$_{\pm0.0005}$ \\
        & HGEN feature drop+edge drop+regularizer & 0.912$_{\pm0.0142}$ & 0.983$_{\pm0.0014}$ \\ \hline
DBLP    & HGEN feature drop+regularizer           & \textbf{0.936}$_{\pm0.0021}$ & \textbf{0.989}$_{\pm0.0015}$ \\
        & HGEN feature drop                       & 0.933$_{\pm0.0030}$ & 0.988$_{\pm0.0004}$ \\
        & HGEN edge drop+regularizer              & 0.916$_{\pm0.0220}$ & 0.985$_{\pm0.0060}$ \\
        & HGEN feature drop+edge drop+regularizer & 0.921$_{\pm0.0168}$ & 0.987$_{\pm0.0044}$ \\ \hline
Business & HGEN feature drop+regularizer           & \textbf{0.732}$_{\pm0.0019}$ & \textbf{0.787}$_{\pm0.0018}$ \\
        & HGEN feature drop                       & 0.725$_{\pm0.0043}$ & 0.787$_{\pm0.0021}$ \\
        & HGEN edge drop+regularizer              & 0.724$_{\pm0.0045}$ & 0.788$_{\pm0.0032}$ \\
        & HGEN feature drop+edge drop+regularizer & 0.723$_{\pm0.0059}$ & 0.785$_{\pm0.0047}$ \\ \hline
Urban   & HGEN feature drop+regularizer           & \textbf{0.591}$_{\pm0.0142}$ & \textbf{0.850}$_{\pm0.0115}$ \\
        & HGEN feature drop                       & 0.553$_{\pm0.0117}$ & 0.837$_{\pm0.0122}$ \\
        & HGEN edge drop+regularizer              & 0.537$_{\pm0.0185}$ & 0.842$_{\pm0.0083}$ \\
        & HGEN feature drop+edge drop+regularizer & 0.548$_{\pm0.0348}$ & 0.845$_{\pm0.0059}$ \\ \hline
\end{tabular}%
}
\caption{Ablation study results \textit{w.r.t.} regularizer, feature Dropout, edge Dropout (using GraphSAGE as base learners). Node Dropout was left out due to its significant inferior performance.}
\label{tab:ablation study}
\end{table}


\subsubsection{Ablation Study on Augmentations \&  Regularizer.}
Table~\ref{tab:ablation study} reports the results of \method\ using different augmentations (feature/edge dropout), combined with the regularizer across various datasets. Feature dropping diversifies individual GNNs by enabling a broader range of learning process, which improves the models' capacity for generalization. The regularizer ensures predictive consistency by enforcing constraints which further enhances model performance. These combined effects make the framework particularly suitable for handling heterogeneous graphs. We observed that the Urban dataset benefits significantly from the regularizer and feature dropping, showcasing their utility in domains characterized by high heterogeneity and noise.
\begin{figure}
    \centering
    \includegraphics[width=0.8\linewidth]{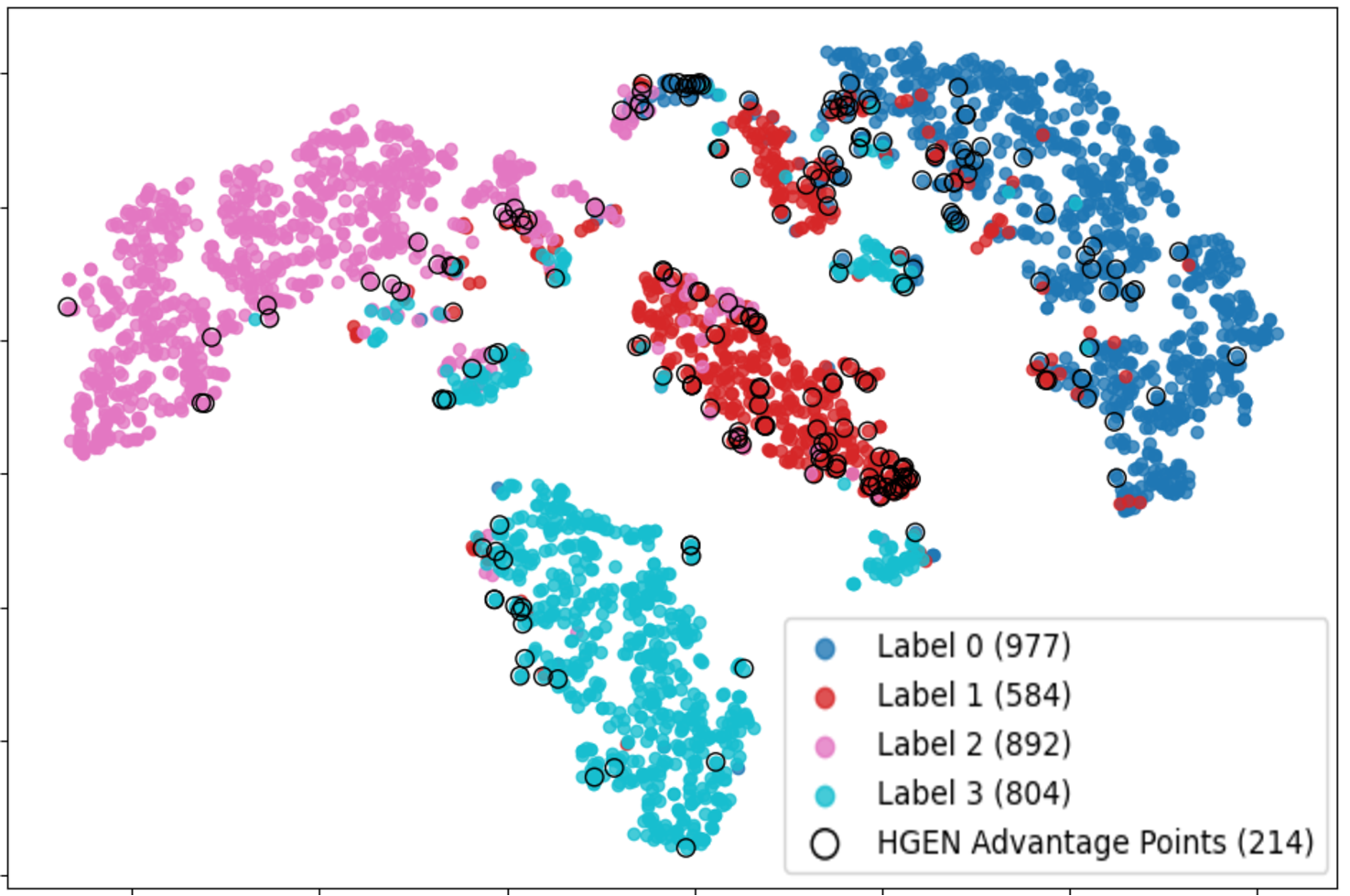}
    \caption{Case study on the DBLP dataset explaining why and how \method\ outperforms HAN using $t$-visualization. Points represent samples in the DBLP dataset, colored-coded based on their ground-truth labels. The circled points are correctly classified by \method\, but misclassified by HAN. There are 214 circled points. }
    \label{fig:TSNE-DBLP}
\end{figure}


\subsection{Case Study Analysis}
To demonstrate why and how \method\ outperforms baselines, we carry out case studies on each benchmark dataset to compare samples on which \method\ makes correct classification whereas rivals (\textit{i.e.} HAN~\cite{han2019}) make mistakes, and report the case study on DBLP in Figure~\ref{fig:TSNE-DBLP} (case studies of rest datasets are reported in Supplement C). For each dataset, points represent samples which are color-coded based on their ground-truth class. Circled points are correctly classified by \method\ but misclassified by HAN.

Taking DBLP dataset in Figure~\ref{fig:TSNE-DBLP} as an example, it can be observed that the circled points are mostly around the edges of the clusters, meaning boundary or difficult cases for separation. The proposed \method\ can correctly identify boundary points for each class that the HAN method incorrectly classify. Similar phenomena can be observed from other four datasets. In fact, this is one of the frequently observed advantages that are naturally brought about by ensemble learning~\cite{Ross2020Ensembles,Polikar2006ensemble}.


\section{Conclusion}

This paper proposed \method, a novel ensemble learning framework for heterogeneous graphs. Unlike existing ensemble methods that focus mainly on homogeneous graphs, \method\ addresses the unique challenges posed by graph heterogeneity, including various node and edge types, and the need for accurate and diverse base learners. By leveraging a regularized allele GNN framework, \method\ enhances generalization through feature dropping techniques, promoting diversity among base learners. The residual-attention mechanism further enables adaptive ensemble weighting, ensuring improved predictive performance through dynamic aggregation of allele GNNs.
%


\appendix

\section*{Ethical Statement}

This is basic ML research and entails no ethical issues.

\section*{Acknowledgments}
This work has been supported in part by the National Science Foundation (NSF) under
Grant Nos\, IIS-2236578, IIS-2236579, IIS-2302786, IIS-2441449, IOS-2430224, and IOS-2446522,
the Science Center for Marine Fisheries (SCeMFiS), and the Commonwealth Cyber Initiative (CCI).


\bibliographystyle{named}
\bibliography{ijcai25}

\end{document}